\documentclass[letterpaper]{article} 
\usepackage[preprint]{aaai2027}  
\usepackage[hyphens]{url}  
\usepackage{graphicx} 
\usepackage{natbib}  
\usepackage{caption} 
\usepackage{algorithm}
\usepackage{algorithmic}
\usepackage{amsmath}
\usepackage{multirow}
\usepackage{amssymb}

\usepackage{newfloat}
\usepackage{listings}
\DeclareCaptionStyle{ruled}{labelfont=normalfont,labelsep=colon,strut=off} 
\floatstyle{ruled}
\newfloat{listing}{tb}{lst}{}
\floatname{listing}{Listing}

\usepackage{booktabs}

\title{Efficient Video Dataset Distillation via Cluster-Guided Prototype Blending}
\author{
Chongle Ren\textsuperscript{\rm 1},
Guang Li\textsuperscript{\rm 1}\thanks{Correspondence to Guang Li <guang@lmd.ist.hokudai.ac.jp>},
Wenbo Huang\textsuperscript{\rm 2},
Naoki Saito\textsuperscript{\rm 1},
Takahiro Ogawa\textsuperscript{\rm 1},
Miki Haseyama\textsuperscript{\rm 1}
}

\affiliations{
\textsuperscript{\rm 1}Hokkaido University \qquad\qquad
\textsuperscript{\rm 2}Southeast University
}

\begin{document}

\maketitle

\begin{abstract}
Video dataset distillation aims to compress a large video dataset into a compact surrogate set that preserves its training utility. Most existing approaches synthesize condensed videos through iterative optimization, whose cost is amplified by the temporal dimension. Rather than further reducing the number of optimized variables, we investigate whether effective distilled videos can be constructed without gradient-based optimization of the stored videos. Such a construction-based approach must address three challenges: selecting informative temporal segments, covering diverse intra-class variations under a limited videos-per-class budget, and increasing the information carried by each stored sample. To this end, we propose \textbf{ProtoBlend}, an efficient \emph{select--allocate--blend} framework. First, teacher-guided temporal clip selection retains a high-confidence segment from each source video. Second, cluster-guided prototype allocation partitions the selected clips in the teacher feature space and assigns one distilled slot to each intra-class cluster. Third, each prototype is blended with an in-cluster anchor, while their teacher predictions are combined using the same coefficient to provide mixture-source supervision. Experiments on four trimmed action-recognition benchmarks demonstrate that ProtoBlend achieves a competitive accuracy--efficiency trade-off without iterative optimization of the distilled videos.
\end{abstract}

\section{Introduction}

The progress of video understanding has been driven by increasingly large training datasets~\cite{tang2025video,kong2025multi,sajedi2024data,ni2022expanding}. Modern action-recognition benchmarks contain hundreds of thousands of clips spanning hundreds of categories, with every sample comprising an entire sequence of frames rather than a single image~\cite{zhu2020comprehensive,kong2022human}. This scale introduces substantial costs throughout the model-development cycle. Datasets must be stored and transferred, while architecture search, hyperparameter tuning, and ablation studies require models to be repeatedly trained on the full collection~\cite{sorscher2022beyond,wang2024internvideo2}. Dataset distillation offers a potential solution by replacing the original dataset with a compact surrogate set that retains its training utility~\cite{wang2018dataset,li2022awesome}.

In the image domain, dataset distillation has developed into a broad family of methods~\cite{yu2023dataset}. Most influential approaches treat synthetic samples as learnable variables and update them through iterative optimization. Early methods match the gradients induced by real and synthetic batches~\cite{zhao2021datasetcondensation,zhao2021dataset}, while later methods align feature distributions~\cite{zhao2023distribution,wang2022cafe,sajedi2023datadam,li2025hdd} or training trajectories~\cite{cazenavette2022dataset,li2023ddpp,li2024iadd,guo2024towards}. A parallel line of work improves scalability by increasing the realism, diversity, and information content of the condensed set~\cite{wang2025edf,ran2026tgdd,sun2024rded,li2026difficulty}. Despite their different objectives, these methods generally require repeated forward and backward passes to optimize the synthetic samples.

Extending this paradigm to videos is considerably more expensive~\cite{yu2023dataset,liu2025evolution}. The temporal dimension multiplies the number of optimized variables by the clip length, and each update must process the entire spatiotemporal volume. Existing video dataset distillation methods therefore focus on reducing the synthetic optimization space. VDSD~\cite{wang2024dancing} disentangles a condensed video into static visual content and auxiliary motion information, reducing the number of directly optimized variables. PRISM~\cite{choi2026prism} instead models appearance and motion jointly and progressively inserts sparse learnable key frames where linear interpolation is insufficient to capture nonlinear dynamics. These approaches provide more efficient video parameterizations, but the condensed videos are still obtained through repeated gradient-based updates.

\begin{figure}[t]
\centering
\includegraphics[width=\columnwidth]{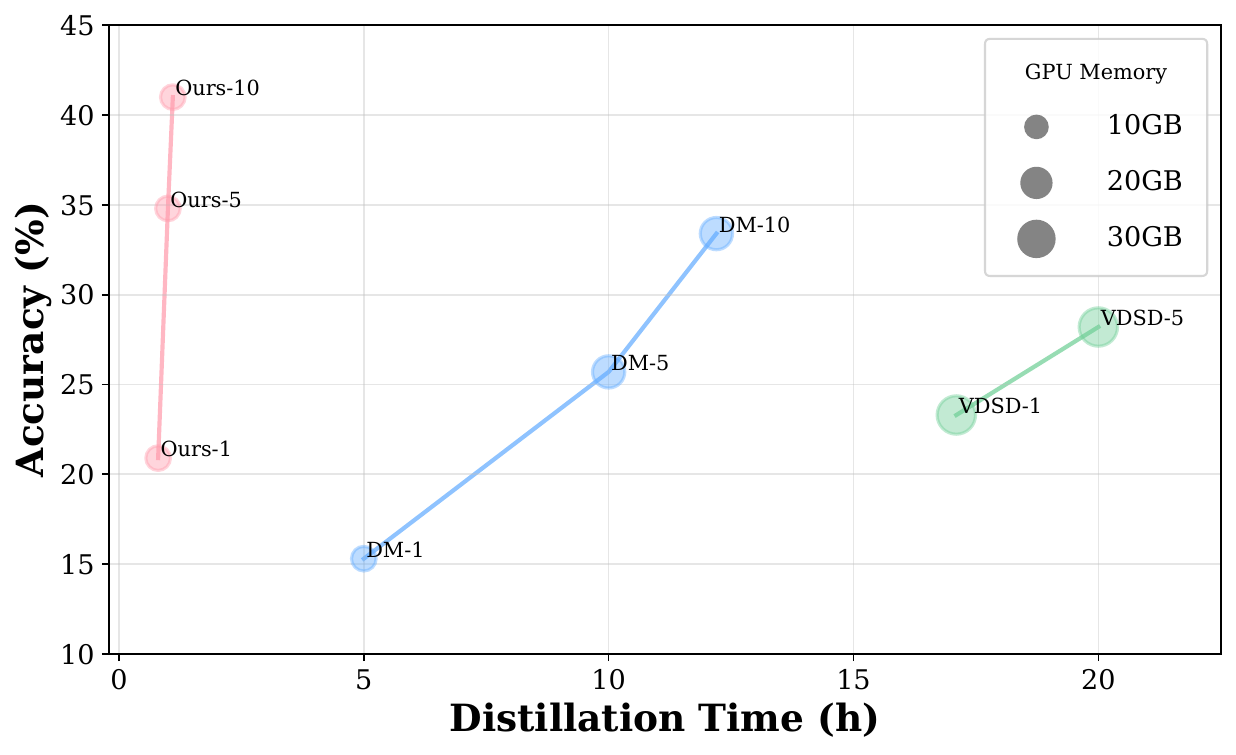}
\caption{
Accuracy versus distilled-set construction time on MiniUCF.
Marker size denotes peak GPU memory. ProtoBlend provides a
favorable trade-off relative to the evaluated optimization-based
baselines.
}
\label{fig:efficiency_overview}
\end{figure}

This motivates the complementary exploration of directly constructing useful distilled videos, without iteratively optimizing the stored videos. Removing synthetic optimization does not by itself produce an effective condensed set. A direct construction strategy must address three challenges. First, raw videos contain redundant or weakly informative temporal segments, making arbitrary clip extraction unreliable. Second, selecting only the most teacher-confident clips may concentrate the limited videos-per-class (VPC) budget on a few easy intra-class regions and overlook other variations. Third, storing one selected clip per slot limits the information contributed by that slot to a single source video. Effective direct construction therefore requires jointly improving temporal clip quality, intra-class coverage, and per-slot information content.

To address these challenges, we propose \textbf{ProtoBlend}, an efficient video dataset distillation framework following a \emph{select--allocate--blend} pipeline. First, Teacher-Guided Temporal Clip Selection (TGS) samples multiple temporal candidates from each source video and retains the clip with the lowest target-class cross-entropy under a frozen teacher. This stage filters out weakly recognizable temporal regions and produces a candidate pool containing one clip per source video. Second, Cluster-Guided Prototype Allocation (CGA) partitions the selected clips within each class using their teacher representations and assigns one distilled slot to each feature-space cluster. Clustering determines where the limited VPC budget is allocated, while teacher confidence determines which clips represent each cluster. Third, Blended Sample Construction combines each selected prototype with an in-cluster anchor through pixel-level interpolation. This allows one stored video to incorporate information from two compatible real clips. To align supervision with the construction process, the teacher predictions of the prototype and anchor are combined using the same blending coefficient.

Once the teacher has been prepared, ProtoBlend constructs the distilled set using forward inference, class-wise clustering, and direct sample blending, without gradient-based updates to the stored videos. As illustrated in Fig.~\ref{fig:efficiency_overview}, this design provides a favorable trade-off between recognition accuracy and construction efficiency. Experiments on MiniUCF, HMDB51, Kinetics-400, and Something-Something V2 further demonstrate competitive performance across different dataset scales and VPC budgets.

Our contributions are summarized as follows:
\begin{itemize}
\item We introduce ProtoBlend, a construction-based video dataset distillation framework that avoids iterative gradient-based optimization of the distilled videos.
\item We develop a \emph{select--allocate--blend} pipeline that addresses temporal clip quality, intra-class budget allocation, and per-slot information content, together with mixture-source soft labels matched to the blended inputs.
\item We evaluate ProtoBlend on four video benchmarks, demonstrating competitive performance across multiple VPC budgets, a favorable accuracy--efficiency trade-off, and transferability to recurrent evaluation architectures on MiniUCF.
\end{itemize}

\begin{figure*}[t]
    \centering
    \includegraphics[width=\textwidth]{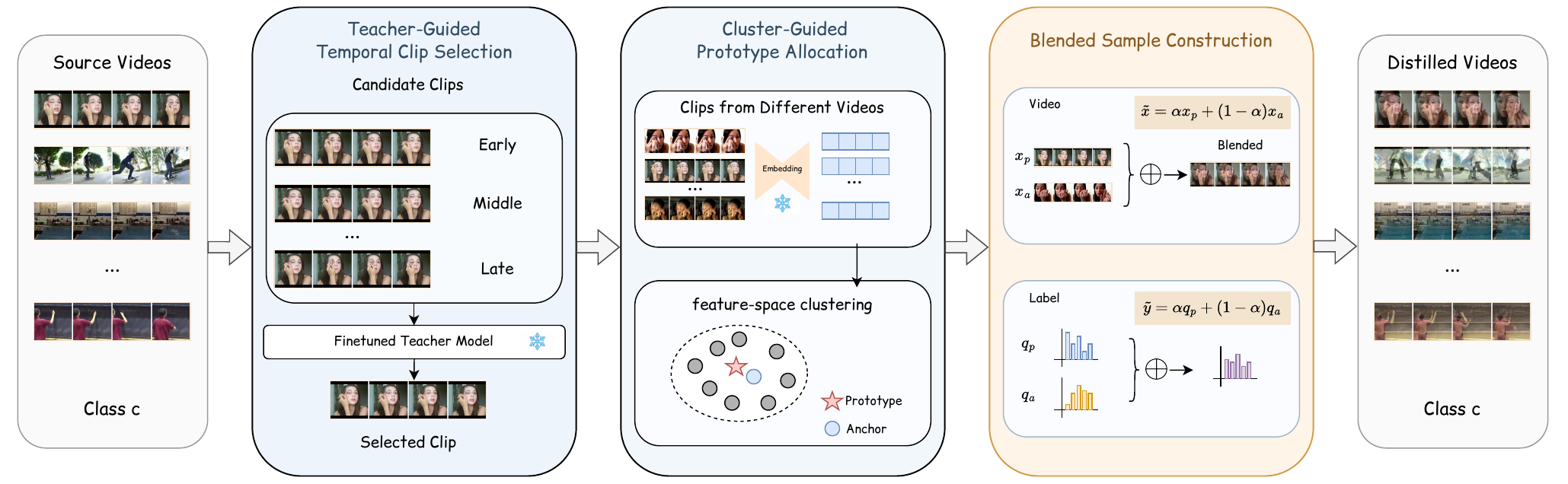}
    \caption{
Overview of ProtoBlend. For each source video, a frozen fine-tuned VideoMAE teacher selects the temporal candidate clip with the lowest target-class cross-entropy loss. The selected clips of each class are then grouped by class-wise $K$-means in the teacher feature space, where the lowest- and second-lowest-loss clips within each cluster are chosen as the prototype and anchor, respectively. Finally, each pair is blended at the pixel level, and their teacher predictions are mixed using the same coefficient to construct the distilled video and its soft label.
}
    \label{fig:protoblend}
\end{figure*}

\section{Related Work}

\subsection{Image Dataset Distillation}

Dataset distillation aims to replace a large training dataset with a compact synthetic set that preserves its training utility. Most existing methods have been developed for image classification and formulate the synthetic samples as learnable variables. Early approaches optimize these samples by matching the gradients induced by real and synthetic batches \cite{zhao2021datasetcondensation,wu2025dataset}. Subsequent methods improve effectiveness and scalability by aligning feature distributions \cite{wang2022cafe,liu2025dataset,cui2025optical,ma2025curriculum}, matching training trajectories \cite{cazenavette2022dataset,guo2024towards,zhong2025towards}, or learning synthetic data through feature regression \cite{zhou2022dataset,cui2025fadrm,wang2025dataset, peng2026clpdd}. More recently, generative dataset distillation has leveraged pretrained generative models to produce diverse and realistic synthetic training samples \cite{li2024generative,ye2025igds,su2024diffusion,zou2025vlcp,cai2026evlf}. These methods have been applied to various downstream tasks, including privacy-preserving learning \cite{li2020soft,li2022compressed,li2023sharing}, fine-grained recognition \cite{ma2026fd2}, and multimodal learning \cite{li2025davdd}. However, applying them directly to videos is considerably more expensive because each sample contains an additional temporal dimension.

\subsection{Video Dataset Distillation}

Video dataset distillation must preserve both spatial appearance and temporal dynamics while controlling the high cost of optimizing video tensors. VDSD \cite{wang2024dancing} addresses this challenge by disentangling each condensed video into static visual content and auxiliary motion information. This parameterization reduces the number of variables that must be directly optimized, making video distillation more tractable than optimizing all frames independently. However, the condensed representation is still learned through iterative gradient-based updates. PRISM \cite{choi2026prism} instead represents each video as a coupled spatiotemporal structure. It initializes a synthetic video with sparse temporal anchors and progressively inserts learnable key frames where linear interpolation is insufficient to capture nonlinear dynamics. This adaptive parameterization allocates optimization capacity to temporally informative regions and improves storage efficiency, but synthesizing the condensed videos still requires repeated forward and backward passes. ProtoBlend explores a complementary construction-based direction that avoids iterative optimization of distilled videos. A fine-tuned, frozen VideoMAE teacher selects high-confidence temporal clips and extracts their features. The videos-per-class budget is then allocated across intra-class clusters, where compatible clips are directly blended to form the distilled set. Unlike VDSD and PRISM, which improve iterative synthesis, ProtoBlend replaces it with teacher-guided selection, allocation, and construction.

\section{Proposed Method}

In this section, we present ProtoBlend, an efficient video dataset distillation method based on a select--allocate--blend pipeline.
Given a training video set
\begin{equation}
\mathcal{D}=\{(x_i,y_i)\}_{i=1}^{N},
\end{equation}
with $C$ classes and a target budget of $K$ videos per class, ProtoBlend
produces a distilled set
\begin{equation}
\tilde{\mathcal{D}}
=
\{(\tilde{x}_{c,k},\tilde{y}_{c,k})\}_{c=1,\,k=1}^{C,\,K},
\end{equation}
containing exactly $C\times K$ stored videos. Each distilled video
$\tilde{x}_{c,k}$ occupies one storage slot but is constructed from two
real clips within the same class. Figure~\ref{fig:protoblend} provides an overview of the framework.

\begin{enumerate}
    \item \textbf{Teacher-Guided Temporal Clip Selection (TGS)} selects the most class-discriminative temporal segment from each source video, improving the quality of the candidate pool.

    \item \textbf{Cluster-Guided Prototype Allocation (CGA)} partitions the selected clips into $K$ teacher-feature modes and assigns one distilled slot to each mode. Within every cluster, teacher confidence is used to select the prototype and anchor, balancing intra-class coverage with discriminative quality.

    \item \textbf{Blended Sample Construction} combines each prototype with its in-mode anchor through pixel-level interpolation. The resulting distilled video therefore incorporates information from two discriminative real clips while occupying only one storage slot. A mixture-source soft label is constructed from the teacher predictions of the same two source clips.
\end{enumerate}

These three stages address complementary aspects of video dataset distillation:
temporal selection improves clip quality, cluster-guided allocation
improves mode coverage, and blending increases the information
density of each stored sample.
 
\subsection{Teacher-Guided Temporal Clip Selection}
 
\label{sec:selection}
Raw training videos often contain temporal segments that are redundant or only weakly discriminative for the target class. For each source video $x_i$, we sample $R$ temporal candidate clips $\mathcal{X}_i=\{x_{i,r}\}_{r=1}^{R}$ and score every candidate by its target-class cross-entropy under the frozen teacher $T$,
\begin{equation}
\ell_{i,r}=\mathrm{CE}\!\big(T(x_{i,r}),\,y_i\big).
\end{equation}
We keep only the most teacher-recognizable clip per source video,
\begin{equation}
x_i^{\star}=\operatorname*{arg\,min}_{x_{i,r}\in\mathcal{X}_i}\ell_{i,r},
\qquad
\ell_i=\min_{r}\ell_{i,r},
\end{equation}
and carry its teacher embedding $z_i=\phi_T(x_i^{\star})$ and score $\ell_i$ to
the next stage. Retaining exactly one clip per source video suppresses temporal noise and prevents a single long video from injecting multiple near-identical segments into the pool. The surviving clips of class $c$ form the candidate pool
\begin{equation}
\mathcal{C}_c=\big\{(x_i^{\star},\,z_i,\,\ell_i)\;:\;y_i=c\big\}.
\end{equation}
The resulting candidate pool suppresses temporally uninformative segments while preserving one discriminative clip from each source video for subsequent allocation.

\subsection{Cluster-Guided Prototype Allocation}
\label{sec:allocation}

Taking the $K$ lowest-loss clips of a class as the distilled set tends to concentrate the limited budget on a few easy intra-class modes, leaving other variations uncovered. We therefore decouple \emph{diversity coverage} from \emph{sample quality}: clustering determines where each distilled slot is allocated, while teacher confidence determines which clips represent each mode.

For each class $c$, we apply $K$-means to the teacher embeddings
$\{z_i:(x_i^{\star},\cdot,\cdot)\in\mathcal{C}_c\}$ and obtain $K$ clusters
$\{\mathcal{G}_{c,k}\}_{k=1}^{K}$. Tying the number of clusters to the VPC budget assigns one stored slot to each feature-space mode, encouraging the distilled set to cover diverse intra-class variations.

Within each cluster, we select clips according to teacher confidence. The prototype is the clip with the lowest target-class cross-entropy,
\begin{equation}
p_{c,k}
=
\operatorname*{arg\,min}_{i\in\mathcal{G}_{c,k}}
\ell_i,
\end{equation}
and the anchor is the second-lowest-loss clip from the same cluster,
\begin{equation}
a_{c,k}
=
\operatorname*{arg\,min}_{i\in\mathcal{G}_{c,k}\setminus\{p_{c,k}\}}
\ell_i.
\end{equation}
Thus, clustering promotes diversity across the distilled set, whereas
loss-based selection preserves discriminative quality within each mode. Since the prototype and anchor originate from the same feature-space cluster, their subsequent interpolation remains local to a coherent intra-class region rather than bridging unrelated modes. If a cluster contains only one clip, we set $a_{c,k}=p_{c,k}$, in which case the following blending step reduces to the original prototype.
 
\subsection{Blended Sample Construction}
\label{sec:blend}
Each distilled slot can store only one video. Simply storing the selected prototype, however, limits the information in that slot to a single real source. We therefore combine the prototype with an in-mode anchor so that one stored video incorporates information from two discriminative clips.

With a fixed blend coefficient $\alpha \in [0.5,1]$, we construct
\begin{equation}
\tilde{x}_{c,k}
=
\alpha x_{p_{c,k}}^{\star}
+
(1-\alpha)x_{a_{c,k}}^{\star},
\end{equation}
a pixel-level, frame-aligned combination of the two clips, which share
the same length and spatial resolution so that no resampling is required. We fix $\alpha$ across the distilled set to maintain a consistent construction rule. Setting $\alpha\geq0.5$ preserves the prototype as the dominant (or, at $\alpha=0.5$, equally weighted) source while allowing the anchor to contribute complementary
information.
 
Supervision is matched to the same mixture. Let $s(x)=\mathrm{softmax}\!\big(T(x)/\tau\big)$ denote the teacher's
temperature-scaled posterior. Rather than querying the teacher on the blended video, whose mixed appearance lies outside the distribution the teacher was trained on, we instead combine the posteriors of the two real sources at the same ratio,
\begin{equation}
\tilde{y}_{c,k}=\alpha\,s\!\big(x_{p_{c,k}}^{\star}\big)+(1-\alpha)\,s\!\big(x_{a_{c,k}}^{\star}\big).
\end{equation}
The mixture-source label is constructed using the same coefficient $\alpha$ as
the input blend, forming a consistently constructed input--target pair
$(\tilde{x}_{c,k},\tilde{y}_{c,k})$ that provides supervision aligned with the
two source clips, without querying the teacher for a potentially unreliable prediction on the out-of-distribution blended video. Since
$s(x_{p_{c,k}}^{\star})$ and $s(x_{a_{c,k}}^{\star})$ are already computed during
selection, constructing $\tilde{y}_{c,k}$ requires no additional teacher
forward pass beyond those used for scoring.
Collecting one pair per cluster across all classes yields the distilled set matching the VPC budget exactly.
\begin{equation}
\tilde{\mathcal{D}}=\big\{(\tilde{x}_{c,k},\tilde{y}_{c,k})\big\}_{c=1,\,k=1}^{C,\,K},
\qquad \big|\tilde{\mathcal{D}}\big|=C\times K,
\end{equation}

\subsection{Training Objective}
\label{sec:objective}
A student classifier $f_\theta$ is trained from random initialization on the
distilled set $\tilde{\mathcal{D}}$. Because every blend combines two clips of the same class, each distilled video carries an unambiguous hard label $c$ alongside the mixture-source soft label $\tilde{y}_{c,k}$, and we
supervise the student with both. For a distilled pair $(\tilde{x},\tilde{y})$ of
class $c$, the hard term is the standard cross-entropy on the student logits,
\begin{equation}
\mathcal{L}_{\mathrm{hard}}=\mathrm{CE}\!\big(f_\theta(\tilde{x}),\,c\big),
\end{equation}
and the soft term is a temperature-scaled distillation loss matching the student
to the stored mixture-source label,
\begin{equation}
\mathcal{L}_{\mathrm{soft}}=\tau^{2}\,\mathrm{KL}\!\big(\tilde{y}\;\big\|\;\mathrm{softmax}(f_\theta(\tilde{x})/\tau)\big),
\end{equation}
using the same temperature $\tau$ that produced $\tilde{y}$; the $\tau^{2}$ factor keeps the gradient magnitude of the soft term comparable to that of the hard term
\begin{equation}
\mathcal{L}=(1-\lambda)\,\mathcal{L}_{\mathrm{hard}}
+\lambda\,\mathcal{L}_{\mathrm{soft}},
\end{equation}
where $\lambda\in[0,1]$ controls their relative contributions. The hard-label term preserves the shared class-level target of the source clips, while the soft-label term transfers finer teacher-derived information matched to the blended input. Once the distilled set and its labels have been constructed, student training requires access to neither the original dataset nor the teacher model, allowing the resulting distilled set to serve as a self-contained and reusable training surrogate. 

\begin{table*}[t]
\centering
\small
\begin{tabular*}{\textwidth}{@{\extracolsep{\fill}}lcccccc}
\toprule
\multirow{2}{*}{Method}
& \multicolumn{3}{c}{MiniUCF}
& \multicolumn{3}{c}{HMDB51} \\
\cmidrule(lr){2-4}\cmidrule(lr){5-7}
& VPC=1 & VPC=5 & VPC=10
& VPC=1 & VPC=5 & VPC=10 \\
\midrule
Random
& $10.3{\pm}1.1$ & $20.7{\pm}0.5$ & $26.0{\pm}0.6$
& $3.8{\pm}0.4$  & $6.5{\pm}0.5$  & $9.0{\pm}0.5$ \\

Herding
& $11.4{\pm}0.3$ & $22.3{\pm}0.1$ & $28.7{\pm}0.9$
& $4.1{\pm}0.3$  & $8.5{\pm}0.5$  & $10.6{\pm}0.2$ \\

K-center
& $10.0{\pm}1.5$ & $19.2{\pm}0.0$ & $26.0{\pm}0.8$
& $3.7{\pm}0.7$  & $6.9{\pm}0.1$  & $7.4{\pm}0.1$ \\
\midrule

DM
& $15.3{\pm}1.1$ & $25.7{\pm}0.2$ & $30.0{\pm}0.6$
& $6.1{\pm}0.2$  & $8.0{\pm}0.2$  & $12.1{\pm}0.4$ \\

FRePo
& $20.3{\pm}0.5$ & $30.2{\pm}1.7$ & --
& $7.2{\pm}0.8$  & $9.6{\pm}0.7$  & -- \\

DM+VDSD
& $17.5{\pm}0.1$ & $27.2{\pm}0.4$ & --
& $6.0{\pm}0.4$  & $8.2{\pm}0.1$  & -- \\

FRePo+VDSD
& $\mathbf{22.0{\pm}1.0}$
& $\underline{31.2{\pm}0.7}$ & --
& $\underline{8.6{\pm}0.5}$
& $10.3{\pm}0.6$ & -- \\

PRISM
& $17.9{\pm}0.3$ & $28.0{\pm}0.1$
& $\underline{31.0{\pm}0.1}$
& $7.5{\pm}0.3$
& $\underline{10.5{\pm}0.4}$
& $\underline{12.8{\pm}0.2}$ \\

\textbf{ProtoBlend}
& $\underline{20.9{\pm}0.1}$
& $\mathbf{34.8{\pm}0.9}$
& $\mathbf{41.0{\pm}0.3}$
& $\mathbf{9.0{\pm}0.1}$
& $\mathbf{12.1{\pm}0.6}$
& $\mathbf{14.2{\pm}0.4}$ \\

\midrule
Full Dataset
& \multicolumn{3}{c}{57.8$\pm$1.1}
& \multicolumn{3}{c}{25.4$\pm$0.2} \\

\bottomrule
\end{tabular*}
\caption{
Comparison on MiniUCF and HMDB51 across different VPC settings. Results are reported as the mean and standard deviation over three runs. Full Dataset denotes training on the complete real training set, and ``--'' indicates an unavailable result. The best and second-best results in each setting are shown in bold and underlined, respectively.
}
\label{tab:main}
\end{table*}

\section{Experiments}

\subsection{Experimental Setup}

\noindent\textbf{Datasets.} We evaluate ProtoBlend on four action-recognition benchmarks spanning different dataset scales: MiniUCF, HMDB51~\cite{kuehne2011hmdb}, Kinetics-400~\cite{carreira2017quo}, and Something-Something V2~\cite{goyal2017something}. Following prior work~\cite{wang2024dancing}, MiniUCF is constructed by selecting the 50 most frequent action classes from UCF101~\cite{soomro2012ucf101}, which contains 13,320 videos from 101 classes. HMDB51 contains 6,849 video clips from 51 action classes. For large-scale evaluation, Kinetics-400 covers 400 human action classes, while Something-Something V2 contains 174 motion-centric action classes.

\noindent\textbf{Baselines.} We compare ProtoBlend with three coreset selection baselines: Random, Herding~\cite{welling2009herding}, and K-center~\cite{sener2017active}. We also include representative dataset distillation methods, including the image-based DM~\cite{zhao2023distribution} and FRePo~\cite{zhou2022dataset}, the VDSD-based variants DM+VDSD and FRePo+VDSD~\cite{wang2024dancing}, and the video-specific PRISM~\cite{choi2026prism}. The VDSD-based variants combine the corresponding image distillation method with VDSD's static--dynamic video representation. All methods are compared under the same VPC budgets and student-evaluation protocol.

\noindent\textbf{Implementation details.} We use a VideoMAE teacher~\cite{tong2022videomae} fine-tuned separately on each source dataset for candidate scoring, feature extraction, and soft-label generation. Each temporal candidate contains 16 frames sampled with a temporal stride of 4. Unless otherwise specified, we set the blend coefficient to $\alpha=0.6$, the soft-label loss weight to $\lambda=0.6$, and the temperature to $\tau=4$. Complete implementation and training hyperparameters are provided in the supplementary material.

\noindent\textbf{Evaluation protocol.} Following prior video dataset distillation studies, we use a four-layer ConvNet3D as the default student architecture. Each student is trained from random initialization, and every experiment is repeated independently three times; we report the mean and standard deviation. Top-1 classification accuracy is used for MiniUCF and HMDB51, while top-5 accuracy is reported for Kinetics-400 and Something-Something V2. To evaluate cross-architecture transfer on MiniUCF at VPC=1, we additionally train CNN+GRU and CNN+LSTM students on the same distilled sets. Within each evaluation setting, all methods use the same student architecture, training schedule, and data augmentation to ensure comparability.

\subsection{Main Results}
\label{EXP:main}

\noindent \textbf{Results on MiniUCF and HMDB51.}
ProtoBlend's clearest advantage on MiniUCF emerges once more than one video per class is available (Table~\ref{tab:main}). At VPC=1, it ranks second, trailing FRePo+VDSD by 1.1 percentage points while remaining ahead of the other baselines. This setting leaves little room for cluster-guided allocation: a single prototype--anchor pair must represent the entire class, regardless of its internal diversity. At VPC=5, ProtoBlend moves ahead of FRePo+VDSD by 3.6 points, and its advantage over PRISM reaches 10.0 points at VPC=10, where several optimization-based baselines do not provide results. With multiple slots, the allocation stage can distribute pairs across distinct feature regions rather than concentrating all capacity on one dominant mode.

HMDB51 shows the same budget-dependent benefit more consistently. ProtoBlend ranks first at all three budgets, but its lead is only 0.4 points at VPC=1 and grows beyond one point at VPC=5 and VPC=10. The improvement at larger budgets is especially relevant because the temporal selection rule remains unchanged; the additional capacity is used primarily to broaden intra-class coverage. Across the two benchmarks, ProtoBlend is competitive rather than uniformly dominant under extreme compression, while its advantage becomes clearer when the budget allows the proposed cluster structure to take effect.

\begin{table}[t]
\centering
\resizebox{\columnwidth}{!}{
\begin{tabular}{lcccc}
\toprule
\multirow{2}{*}{Method}
& \multicolumn{2}{c}{SSv2}
& \multicolumn{2}{c}{Kinetics-400} \\
\cmidrule(lr){2-3}\cmidrule(lr){4-5}
& VPC=1 & VPC=5 & VPC=1 & VPC=5 \\
\midrule
Random
& $3.1{\pm}0.1$ & $3.6{\pm}0.1$
& $3.0{\pm}0.2$ & $5.5{\pm}0.2$ \\

Herding
& $2.8{\pm}0.1$ & $3.6{\pm}0.1$
& $3.3{\pm}0.1$ & $6.3{\pm}0.2$ \\

K-center
& $2.6{\pm}0.2$
& $\mathbf{4.5{\pm}0.1}$
& $3.1{\pm}0.1$
& $6.2{\pm}0.2$ \\
\midrule

DM
& $3.6{\pm}0.0$
& $4.1{\pm}0.0$
& $6.3{\pm}0.0$
& $\mathbf{9.1{\pm}0.9}$ \\

DM+VDSD
& $3.8{\pm}0.1$
& $4.0{\pm}0.1$
& $6.3{\pm}0.2$
& $7.0{\pm}0.1$ \\

PRISM
& $\underline{3.9{\pm}0.2}$
& $4.1{\pm}0.1$
& $\underline{7.1{\pm}0.1}$
& $\underline{8.1{\pm}0.1}$ \\

\textbf{ProtoBlend}
& $\mathbf{4.2{\pm}0.2}$
& $\underline{4.3{\pm}0.0}$
& $\mathbf{7.2{\pm}0.1}$
& $8.0{\pm}0.3$ \\

\midrule
Full Dataset
& \multicolumn{2}{c}{34.6$\pm$0.5}
& \multicolumn{2}{c}{29.0$\pm$0.6} \\

\bottomrule
\end{tabular}}
\caption{
Comparison on SSv2 and Kinetics-400 across different VPC settings.
}
\label{tab:large_scale}
\end{table}

\noindent \textbf{Results on SSv2 and Kinetics-400.}
The scaling behavior changes on the larger SSv2 and Kinetics-400 benchmarks (Table~\ref{tab:large_scale}). On SSv2, ProtoBlend has a 0.3-point edge over PRISM at VPC=1, within the observed run-to-run variation, but increasing the budget adds only 0.1 points and leaves it slightly behind K-center at VPC=5. SSv2 classes are often distinguished by fine-grained motion rather than appearance alone. Clustering teacher features can diversify the selected content, but frame-wise blending does not align the motion phases of the prototype and anchor, limiting the value of additional pairs.

Kinetics-400 presents a related limitation in a more heterogeneous setting. ProtoBlend is nearly tied with PRISM at VPC=1 but trails DM by 1.1 points at VPC=5, again showing limited returns from the larger budget. A fixed prototype--anchor construction is effective for selecting a compact representative, yet becomes less flexible as intra-class variation in scene, appearance, and motion increases. More explicit temporal alignment and class-adaptive prototype or budget allocation are therefore promising extensions for large-scale video distillation.

\subsection{Efficiency Analysis}

\begin{figure}[t]
    \centering
    \includegraphics[width=\columnwidth]{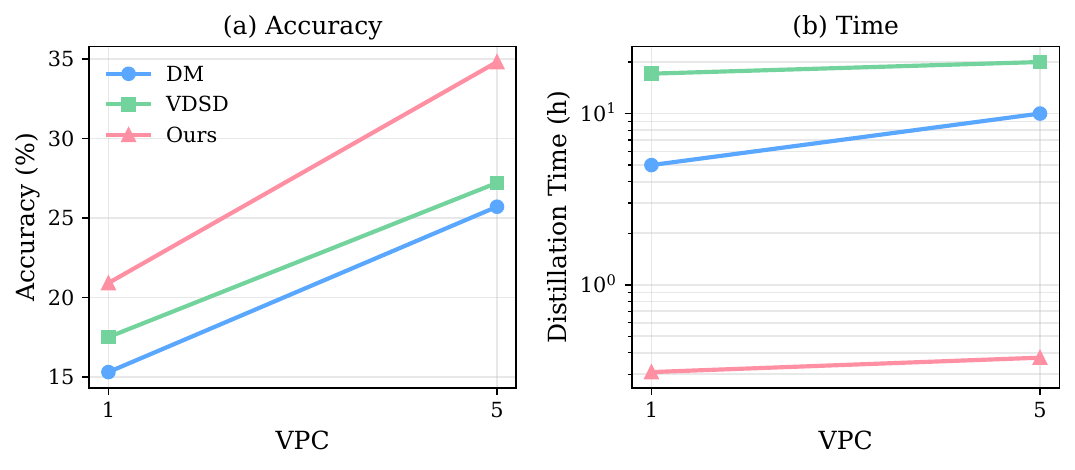}
    \caption{Accuracy and distillation time on MiniUCF at VPC=1 and VPC=5. ProtoBlend achieves higher accuracy with substantially lower distillation cost.}
    \label{fig:efficiency}
\end{figure}

ProtoBlend shifts the main construction cost from iterative video optimization to a single teacher-inference stage. In Fig.~\ref{fig:efficiency}, its post-teacher construction time remains below one hour at both VPC=1 and VPC=5, while its recognition accuracy exceeds that of DM and DM+VDSD. The optimization-based methods repeatedly update every condensed video through forward and backward passes, causing their computational cost to grow with the number of distilled videos.

ProtoBlend instead caches the teacher features, confidence scores, and soft predictions for all temporal candidates. Increasing VPC then requires only class-wise clustering, prototype--anchor selection, and pixel-level blending, without restarting synthetic-data optimization. Its runtime therefore changes only slightly from VPC=1 to VPC=5, while the gap to the optimization-based baselines widens. This weak dependence on the target budget accounts for ProtoBlend's favorable accuracy--efficiency trade-off and allows the cached teacher outputs to be reused when constructing distilled sets at different compression levels.

\subsection{Cross-Architecture Transfer on MiniUCF}

We evaluate the same MiniUCF distilled sets at VPC=1 with CNN+GRU and CNN+LSTM to test whether their effectiveness depends on the default ConvNet3D evaluator. ProtoBlend ranks first with all three architectures, while its accuracy varies by only 1.1 percentage points across them. By comparison, DM and DM+VDSD lose more than five points when moving from ConvNet3D to either recurrent model. PRISM is more stable but remains less accurate, with ProtoBlend leading it by 1.0 point on CNN+GRU and 2.8 points on CNN+LSTM. ProtoBlend constructs its videos from frozen teacher features and class posteriors without using gradients from the evaluation architecture. The retained visual and temporal cues are therefore not tailored to ConvNet3D and remain accessible to recurrent temporal models, suggesting that the accuracy gain is not specific to the ConvNet3D evaluator.

\begin{table}[t]
\centering
\renewcommand{\arraystretch}{1.15}
\resizebox{\columnwidth}{!}{
\begin{tabular}{lccc}
\toprule
\multirow{2}{*}{Method} 
& \multicolumn{3}{c}{Evaluation Model} \\
\cmidrule(lr){2-4}
& ConvNet3D & CNN+GRU & CNN+LSTM \\
\midrule
DM              & $15.3 \pm 1.1$ & $9.9 \pm 0.7$  & $9.2 \pm 0.3$  \\
DM+VDSD      & $17.5 \pm 0.1$ & $12.0 \pm 0.7$ & $10.3 \pm 0.2$ \\
PRISM           & $17.9 \pm 0.3$ & $18.9 \pm 0.8$ & $18.2 \pm 1.3$ \\
\textbf{ProtoBlend}   & $\mathbf{20.9 \pm 0.1}$ 
                & $\mathbf{19.9 \pm 0.1}$ 
                & $\mathbf{21.0 \pm 0.1}$ \\
\bottomrule
\end{tabular}
}
\caption{Cross-architecture evaluation on MiniUCF (VPC=1). 
All methods are evaluated using ConvNet3D, CNN+GRU, and CNN+LSTM. 
Higher values are better.}
\label{tab:cross_arche}
\end{table}

\subsection{Ablation Study}
\label{sec:ablation}

Temporal selection is the dominant contributor in Table~\ref{tab:ablation}. TGS improves accuracy by 8.2 and 8.9 percentage points at VPC=1 and VPC=5, respectively, confirming that the quality of the temporal candidates is more important than the subsequent construction choices. Filtering weakly recognizable segments gives the allocation and blending stages a more discriminative candidate pool. CGA has a different, budget-dependent role: it has little effect when VPC=1 forces all samples into a single cluster, but contributes a further 4.2-point gain at VPC=5. The contrast isolates CGA as a mechanism for distributing additional slots across intra-class modes rather than improving a single representative.

Blending cannot be assessed independently of its supervision. With hard labels, it adds 1.4 points at VPC=1 but loses 1.5 points at VPC=5. A blended clip can enrich a single representative, whereas interpolating already-diverse cluster prototypes may also introduce visual and supervisory ambiguity. Matching the label to the two mixture sources resolves this conflict, improving the blended construction by 1.6 and 2.5 points at the two budgets and producing the best configuration. The final gain therefore comes from coupling prototype--anchor blending with mixture-consistent supervision, rather than treating blending as a standalone augmentation.

\begin{table}[t]
\centering
\small
\renewcommand{\arraystretch}{1.20}
\setlength{\tabcolsep}{4.8pt}
\begin{tabular}{cccccc}
\toprule
\multicolumn{4}{c}{Components} &
\multicolumn{2}{c}{MiniUCF} \\
\cmidrule(lr){1-4}\cmidrule(lr){5-6}
TGS & CGA & Blend & Soft & VPC=1 & VPC=5 \\
\midrule
--         & --         & --         & -- &
$10.3{\pm}1.1$ & $20.7{\pm}0.5$ \\

\checkmark & --         & --         & -- &
$18.5{\pm}0.2$ & $29.6{\pm}0.4$ \\

\checkmark & \checkmark & --         & -- &
$17.9{\pm}0.9$ & $33.8{\pm}0.3$ \\

\checkmark & \checkmark & \checkmark & -- &
$\mathrm{19.3{\pm}0.9}$ & $\mathrm{32.3{\pm}0.4}$ \\

\checkmark & \checkmark & \checkmark & \checkmark &
$\mathbf{20.9{\pm}0.1}$ & $\mathbf{34.8{\pm}0.9}$ \\
\bottomrule
\end{tabular}
\caption{
Ablation study on MiniUCF. TGS denotes Teacher-Guided Temporal Clip Selection, CGA denotes Cluster-Guided Prototype Allocation, Blend denotes the prototype--anchor blending, and Soft denotes mixture-source soft-label supervision. 
}
\label{tab:ablation}
\end{table}

\subsection{Parameter Sensitivity Analysis}

\begin{figure}[t]
    \centering
    \includegraphics[width=\columnwidth]
    {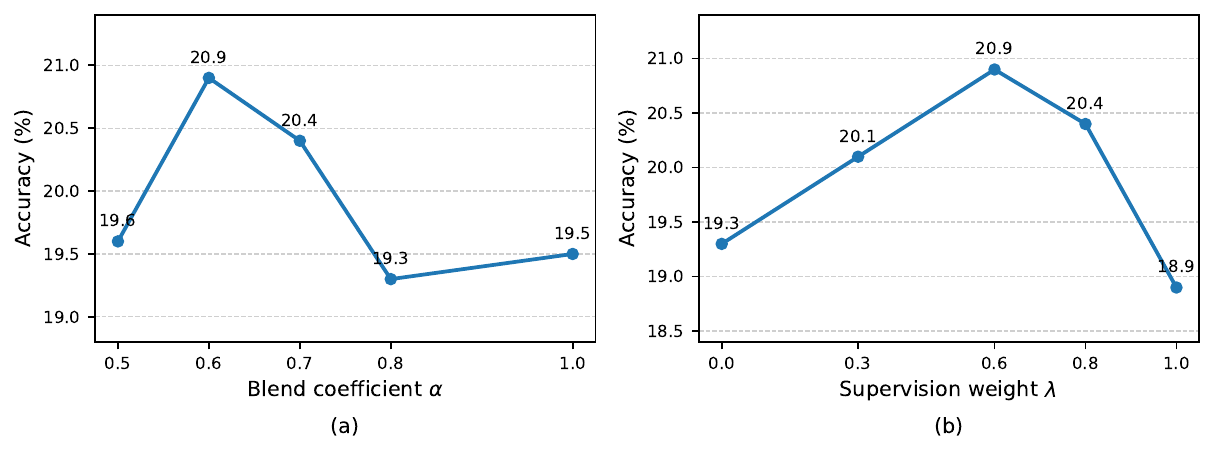}
    \caption{
    Sensitivity to the blend coefficient $\alpha$ and supervision weight
    $\lambda$ on MiniUCF at VPC=1. 
    }
    \label{fig:parameter_sensitivity}
\end{figure}

Both hyperparameters favor balanced mixtures rather than degenerate single-source settings (Fig.~\ref{fig:parameter_sensitivity}). The best blend coefficient is $\alpha=0.6$, where the prototype remains dominant, and the anchor supplies complementary content. Moving $\alpha$ toward 1 gradually removes the anchor contribution and reduces blending to prototype selection, while assigning the anchor equal influence can dilute the discriminative content of the prototype. The interior optimum therefore agrees with the intended division of roles between the two selected clips.

The supervision weight follows a similar principle, with $\lambda=0.6$ providing the best balance. Hard-label supervision alone ignores the uncertainty introduced by combining two inputs, whereas relying entirely on the mixture-source posterior weakens the explicit class-level target. Their combination retains class identity while adapting the prediction target to the blended content. Accuracy changes smoothly around the selected values rather than collapsing outside a narrow optimum, so we use $\alpha=0.6$ and $\lambda=0.6$ throughout the remaining experiments.

\begin{table}[t]
\centering
\small
\renewcommand{\arraystretch}{1.20}
\setlength{\tabcolsep}{4.5pt}
\begin{tabular}{llcc}
\toprule
Construction & Supervision & VPC=1 & VPC=5 \\
\midrule
\multirow{2}{*}{Prototype-only}
    & Hard & $17.9{\pm}0.9$ & $33.8{\pm}0.3$ \\
    & Hard+Soft & $19.5{\pm}0.3$  & $32.7{\pm}0.2$  \\
\cmidrule{1-4}
\multirow{2}{*}{Blend}
    & Hard & $19.3{\pm}0.9$ & $32.3{\pm}0.4$ \\
    & Hard+Soft & $\mathbf{20.9{\pm}0.1}$ & $\mathbf{34.8{\pm}0.9}$ \\
\bottomrule
\end{tabular}
\caption{
Comparison of construction and supervision strategies on MiniUCF. Both Prototype-only and Blend include TGS and CGA. Hard+Soft denotes joint hard-label and mixture-source soft-label supervision.
}
\label{tab:disentangle}
\end{table}

\subsection{Component Interaction}

Blending and soft-label supervision are not independently additive (Table~\ref{tab:disentangle}). With hard labels, blending adds 1.4 percentage points at VPC=1 but loses 1.5 points at VPC=5. Similarly, adding soft-label supervision to prototype-only samples improves VPC=1 by 1.5 points but reduces VPC=5 by 1.1 points. When combined, however, soft-label supervision improves the hard-label blending baseline by 1.6 and 2.5 points at VPC=1 and VPC=5, respectively, yielding the best configuration in both settings. A hard label ignores the relative contributions of the blended sources, whereas the mixture-source posterior changes consistently with the visual input. Thus, the benefit of soft supervision arises from aligning the target with the construction process rather than acting as an independent regularizer.

\begin{figure}[t]
    \centering
    \includegraphics[width=\columnwidth]
    {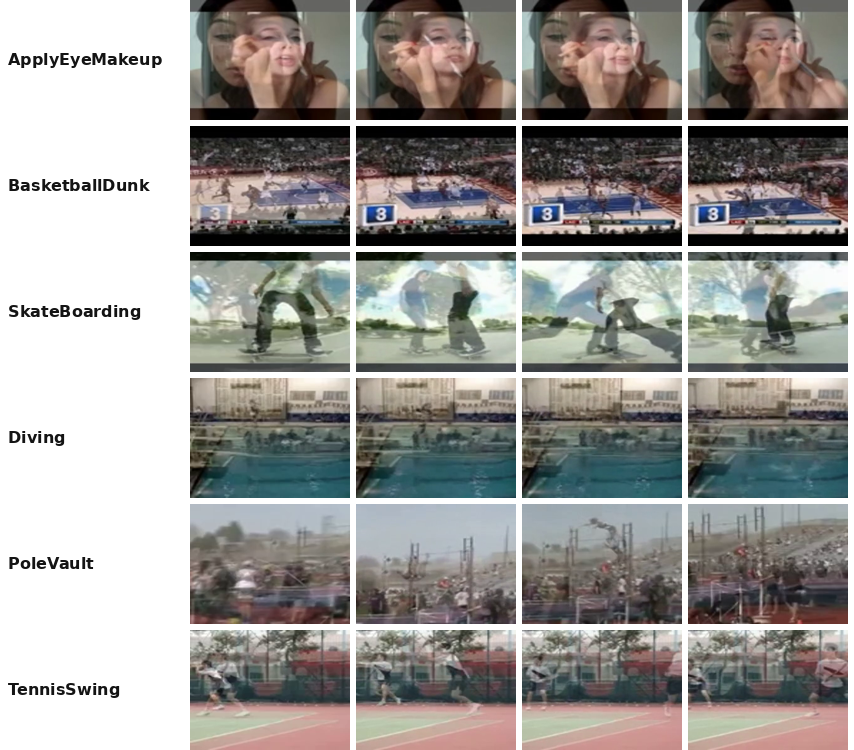}
    \caption{
    Examples of distilled videos by ProtoBlend.
    }
    \label{fig:qualitative}
\end{figure}

\subsection{Qualitative Visualization}

Figure~\ref{fig:qualitative} makes the visual trade-off of prototype--anchor blending explicit. In the appearance-oriented ApplyEyeMakeup example, the face, hand, and manipulated facial region remain identifiable. Motion-oriented examples retain both scene context and ordered changes in body pose, including the approach and execution phases of Diving and PoleVault. Pixel-level superposition introduces visible ghosting in some frames, but it does not erase the cues that distinguish the displayed actions. ProtoBlend therefore prioritizes complementary class and motion information over photorealism, which is appropriate for constructing compact training surrogates. Additional classes and temporal samples are included in the supplementary material.

\section{Conclusion}
In this paper, we presented ProtoBlend, an efficient video dataset distillation framework that replaces iterative optimization with a select--allocate--blend pipeline. ProtoBlend uses a frozen teacher to select high-confidence temporal clips, allocate distilled slots across intra-class feature clusters, and blend compatible prototype--anchor pairs with mixture-source soft labels. Experiments on four trimmed action-recognition benchmarks demonstrate a competitive accuracy--efficiency trade-off across the evaluated VPC settings, along with lower post-teacher construction cost than the evaluated optimization-based baselines, and transferability to recurrent architectures on MiniUCF.

\bibliography{aaai2027}

\newpage

\section{Training Details}
We use dataset-specific VideoMAE-Base teachers for candidate scoring, feature
extraction, and soft-label generation. ProtoBlend samples five temporal
candidates from each source video and selects the candidate with the lowest
target-class cross-entropy. Class-wise K-means is then applied to the teacher
embeddings to allocate the VPC budget and select prototype--anchor pairs.
 
\begin{table}[!ht]
\centering
\small
\renewcommand{\arraystretch}{1.12}
\setlength{\tabcolsep}{5pt}
\begin{tabular}{lc}
\toprule
Setting & Value \\
\midrule
Frames per clip & 16 \\
Temporal stride & 4 \\
Temporal candidates per video ($R$) & 5 \\
Teacher inference batch size & 4 \\
Clustering method & K-means \\
Blend coefficient $\alpha$ & 0.6 \\
Soft-label weight $\lambda$ & 0.6 \\
Temperature $\tau$ & 4 \\
Student model & ConvNet3D \\
Student training epochs & 500 \\
Evaluation repeats & 3 \\
Student batch size & 64 \\
Student learning rate & 0.001 \\
Student optimizer & SGD \\
Data augmentation & random horizontal flip \\
\bottomrule
\end{tabular}
\caption{Default implementation settings used in our experiments.}
\label{tab:supp_settings}
\end{table}
 
The distilled videos are exported at $256\times256$ resolution and resized to
$112\times112$ during student training. Unless otherwise specified, all
reported results are averaged over three independent student-training runs.

\section{Effect of Prototype--Anchor Pairing}
\label{sec:pairing_ablation}

We further investigate the effect of prototype--anchor pairing on MiniUCF.
To ensure a fair comparison, all variants use the same budget,
blend coefficient, teacher, and student-training protocol; only the anchor selection strategy is changed.

\begin{table}[t]
\centering
\small
\renewcommand{\arraystretch}{1.15}
\setlength{\tabcolsep}{7pt}
\begin{tabular}{lcc}
\toprule
Pairing Strategy & VPC=1 & VPC=5 \\
\midrule
Prototype only & $17.9{\pm}0.9$ & $33.8{\pm}0.3$ \\
Random same-class pair & $13.5{\pm}0.5$ & $30.3{\pm}0.3$ \\
In-cluster pair (Ours) & $\mathbf{20.9{\pm}0.1}$ & $\mathbf{34.8{\pm}0.9}$ \\
\bottomrule
\end{tabular}
\caption{
Effect of prototype--anchor pairing strategies on MiniUCF.
}
\label{tab:pairing_ablation}
\end{table}

As shown in Table~\ref{tab:pairing_ablation}, random same-class pairing
substantially degrades performance, reducing accuracy from 17.9\% to 13.5\%
at VPC=1 and from 33.8\% to 30.3\% at VPC=5. This result indicates that
sharing the same class label alone does not guarantee compatibility between
two video clips. Randomly paired clips may differ considerably in viewpoint,
appearance, spatial alignment, or motion pattern, making their pixel-level
interpolation less coherent.

In contrast, the proposed in-cluster pairing achieves 20.9\% and 34.8\%
accuracy, outperforming random same-class pairing by 7.4 and 4.5 percentage
points, respectively. It also improves over the prototype-only variant by
3.0 percentage points at VPC=1 and 1.0 percentage point at VPC=5. These
results demonstrate that the gains do not arise from arbitrary sample mixing.

\section{Limitations}
\label{sec:supp_limitations}

ProtoBlend depends on the quality of the dataset-specific teacher, and errors
in teacher scoring or feature extraction may affect the resulting distilled
set. Moreover, frame-aligned pixel interpolation can produce ghosting when
paired clips differ in viewpoint, spatial alignment, or motion phase, especially
for motion-centric actions. Motion-aware pairing and alignment-aware
blending are promising directions for future work.

\section{Qualitative Results}
\label{sec:supp_qualitative}

\begin{figure*}[t]
    \centering
    \includegraphics[width=0.90\textwidth]
    {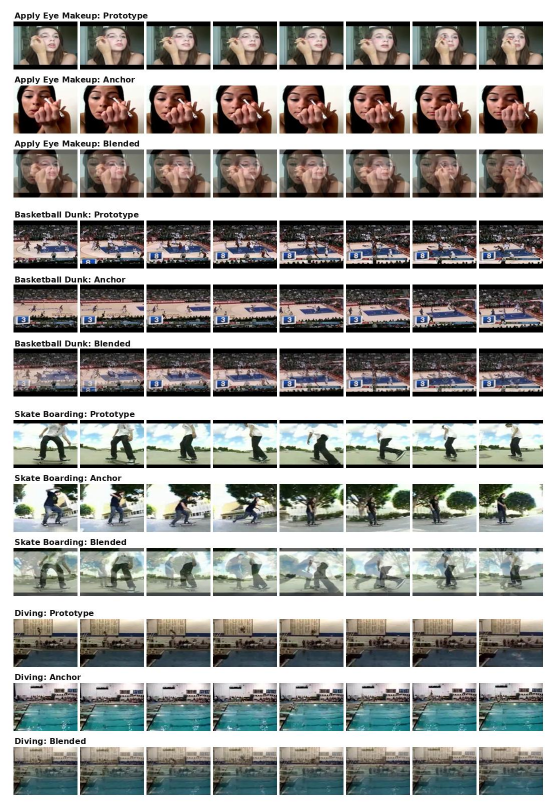}
    \caption{
    Qualitative examples on MiniUCF. For each class, three consecutive rows
    show the prototype, in-cluster anchor, and blended video.
    }
    \label{fig:supp_MiniUCF_qualitative}
\end{figure*}

\begin{figure*}[t]
    \centering
    \includegraphics[width=0.90\textwidth]
    {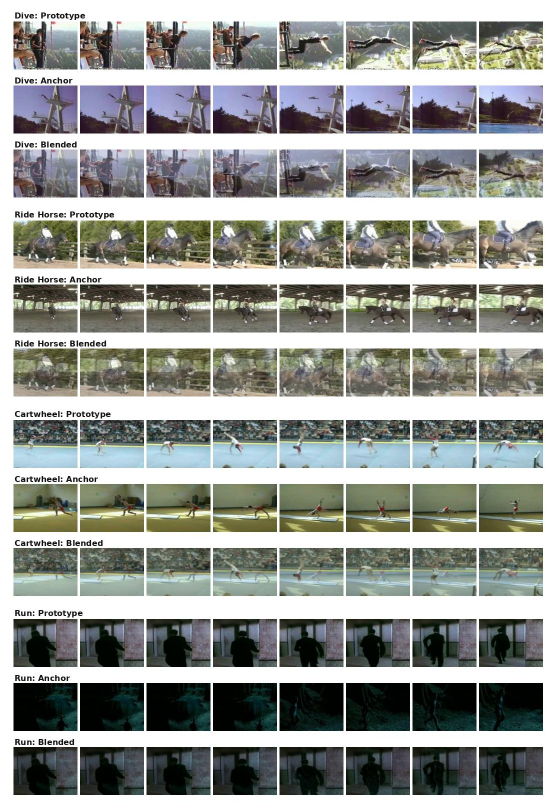}
    \caption{
    Qualitative examples on HMDB51. For each class, three consecutive rows
    show the prototype, in-cluster anchor, and blended video.
    }
    \label{fig:supp_hmdb51_qualitative}
\end{figure*}

\begin{figure*}[t]
    \centering
    \includegraphics[width=0.90\textwidth]
    {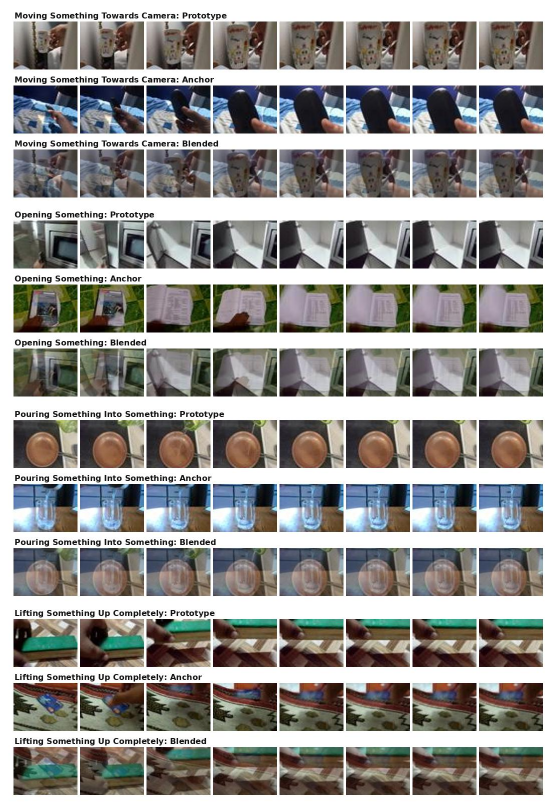}
    \caption{
    Qualitative examples on SSv2. For each class, three consecutive rows show
    the prototype, in-cluster anchor, and blended video.
    }
    \label{fig:supp_ssv2_qualitative}
\end{figure*}

\begin{figure*}[t]
    \centering
    \includegraphics[width=0.90\textwidth]
    {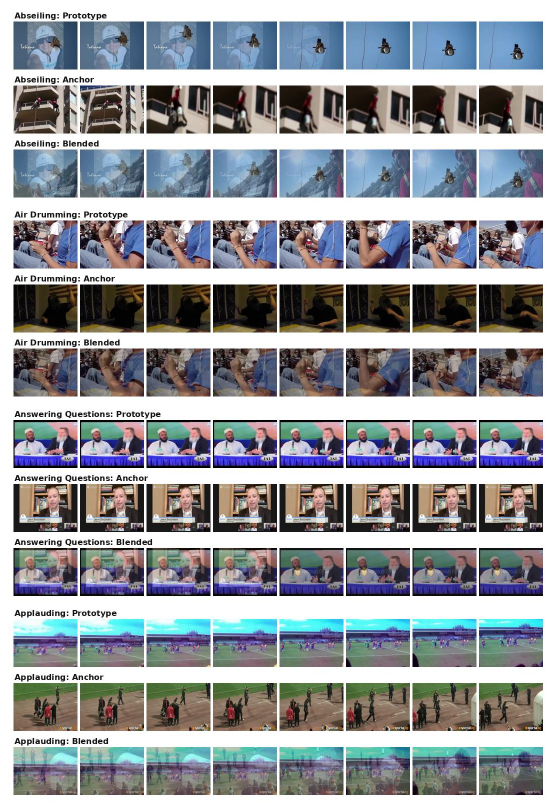}
    \caption{
    Qualitative examples on Kinetics-400. For each class, three consecutive
    rows show the prototype, in-cluster anchor, and blended video.
    }
    \label{fig:supp_kinetics_qualitative}
\end{figure*}

Figures~\ref{fig:supp_MiniUCF_qualitative}--
\ref{fig:supp_kinetics_qualitative} present additional examples on all four
datasets. For each dataset, four representative classes are selected. Each
class is shown using three consecutive rows corresponding to the prototype,
in-cluster anchor, and blended video, with eight temporally ordered frames per row.

The examples show that the prototype and anchor generally share the same
action semantics while differing in appearance, viewpoint, background, or
motion progression. The blended videos preserve the dominant content of the
prototype while incorporating information from the anchor. 

\end{document}